\documentclass[letterpaper, 10 pt, conference]{ieeeconf}  

\IEEEoverridecommandlockouts                              

\usepackage{cite}
\usepackage{amsmath,amssymb,amsfonts}
\usepackage{algorithmic}
\usepackage{graphicx}
\usepackage{subcaption}
\usepackage{textcomp}
\usepackage{xcolor}
\usepackage{comment}
\usepackage[bottom=76pt, top=57pt, left=45pt, right=45 pt ]{geometry}

\definecolor{mygray}{rgb}{0.5, 0.5, 0.5}
\definecolor{orange}{rgb}{1.0, 0.5, 0.2}
\newcommand{\erhan}[1]{\textcolor{black}{#1}}
\newcommand{\hakan}[1]{\textcolor{black}{#1}}

\title{\LARGE \bf
Correspondence learning between morphologically different robots via task demonstrations
}

\author{Hakan Aktas$^{1,2}$, Yukie Nagai$^{2}$, Minoru Asada$^{3,4}$, Erhan Oztop$^{3,5}$, Emre Ugur$^{1}$
\thanks{$^{1}$Hakan Aktas and Emre Ugur are with Computer Engineering Department, Bogazici University
        {\tt hakan.aktas1@boun.edu.tr}
$^{2}$Hakan Aktas and Yukie Nagai are with IRCN, the University of Tokyo
$^{3}$Minoru Asada and Erhan Oztop are with SISREC, Osaka University
$^{4}$Minoru Asada is also affiliated with the International Professional University of Technology in Osaka $^{5}$Erhan Oztop is also affiliated with Computer Engineering Department, Ozyegin University.
}}

\begin{document}

\maketitle
\thispagestyle{empty}
\pagestyle{empty}

\begin{abstract}


We observe a large variety of robots in terms of their bodies, sensors, and actuators. Given the commonalities in the skill sets, teaching each skill to each different robot independently is inefficient and not scalable when the large variety in the robotic landscape is considered. If we can learn the correspondences between the sensorimotor spaces of different robots, we can expect a skill that is learned in one robot can be more directly and easily transferred to other robots. In this paper, we propose a method to learn correspondences \hakan{among two or more robots that may have different morphologies. To be specific, besides robots with similar morphologies with different degrees of freedom, we show that a fixed-based manipulator robot with joint control and a differential drive mobile robot can be addressed within the proposed framework. To set up the correspondence among the robots considered, an initial base task is demonstrated to the robots to achieve the same goal. Then, a common latent representation is learned along with the individual robot policies for achieving the goal.} After the initial learning stage, the observation of a new task execution by one robot becomes sufficient to generate a latent space representation pertaining to the other robots to achieve the same task. We verified our system in a set of experiments where the correspondence between robots is learned (1) when the robots need to follow the same paths to achieve the same task, (2) when the robots need to follow different trajectories to achieve the same task, and (3) when complexities of the required sensorimotor trajectories are different for the robots. We also provide a proof-of-the-concept realization of correspondence learning between a real manipulator robot and a simulated mobile robot.

\end{abstract}


\section{INTRODUCTION}

We observe a large variety of robots in terms of their morphology, sensors, and actuators, such as mobile, aerial, underwater robots, and fixed-based manipulators. While the robots are designed for the environments they operate, there are commonalities in the required skills, such as grasping, and releasing objects, grasping and using tools, and pushing and pulling objects. In general, robots are equipped with these skills by using the Learning-from-demonstration approach with additional controller tuning when needed.  
Given the commonalities in the skill sets, teaching each skill to each different robot independently is not efficient and not scalable when the large variety in the robotic landscape is considered. If we can learn the correspondences between the sensorimotor spaces of different robots, we can expect a skill that is learned in one robot can be more directly and easily transferred to the others.



The learning of sensorimotor correspondences between robots has been  studied in robotics, where emphasis was given to efficiency in the transfer of skills from one robot to the others \cite{taylor2009transfer}. The correspondence between the robots has been established between all pairs of states \cite{taylor2008transferring} either by manually forming a shared feature space \cite{ammar2012reinforcement} or by aligning states using unsupervised manifold alignment \cite{ammar2015unsupervised}. State alignment from local and global perspectives has also been used with a regularized policy update \cite{liu2019state}. Expectation-maximization-based dynamic time warping has been utilized to align the states and subsequently established a shared feature space by employing non-linear embedding functions \cite{gupta2017learning}. Transferring policies using modularity have also been established in various studies \cite{devin2017learning},  with some using hierarchical decoupling between agents  \cite{hejna2020hierarchically, sharma2019third}. Methods that can find  unsupervised action correspondence using unpaired\cite{zhang2020learning} and unaligned \cite{kim2020domain}  have also been introduced. While the previous work learned correspondences at \erhan{individual state level, in our previous work \cite{akbulut2021acnmp}, we proposed to form a common feature space for encoding the correspondence between whole robot trajectories by introducing an additional loss term to enforce the latent representations of the involved robots to be the same. This allowed the skill transfer between two fixed-based manipulators with 3 and 4 degrees of freedom, albeit requiring  extra hyperparameter tuning. In contrast, the currently proposed method is free from this hyper-parameter tuning and is directly applicable to any multiplicity of robots with no scalability issues.} \erhan{Although the impressive developments in the LLM-robotic front \cite{padalkar2023open , bousmalis2023robocat}  may, in the future, render robot-to-robot skill transfer obsolete for commonplace robots, our model offers a zero-shot robot-to-robot skill transfer mechanism with limited resources (computational power, robot data),  which can be applied to any type of robot. }


\begin{figure}[t!]
\centerline{\includegraphics[width=0.9\linewidth]{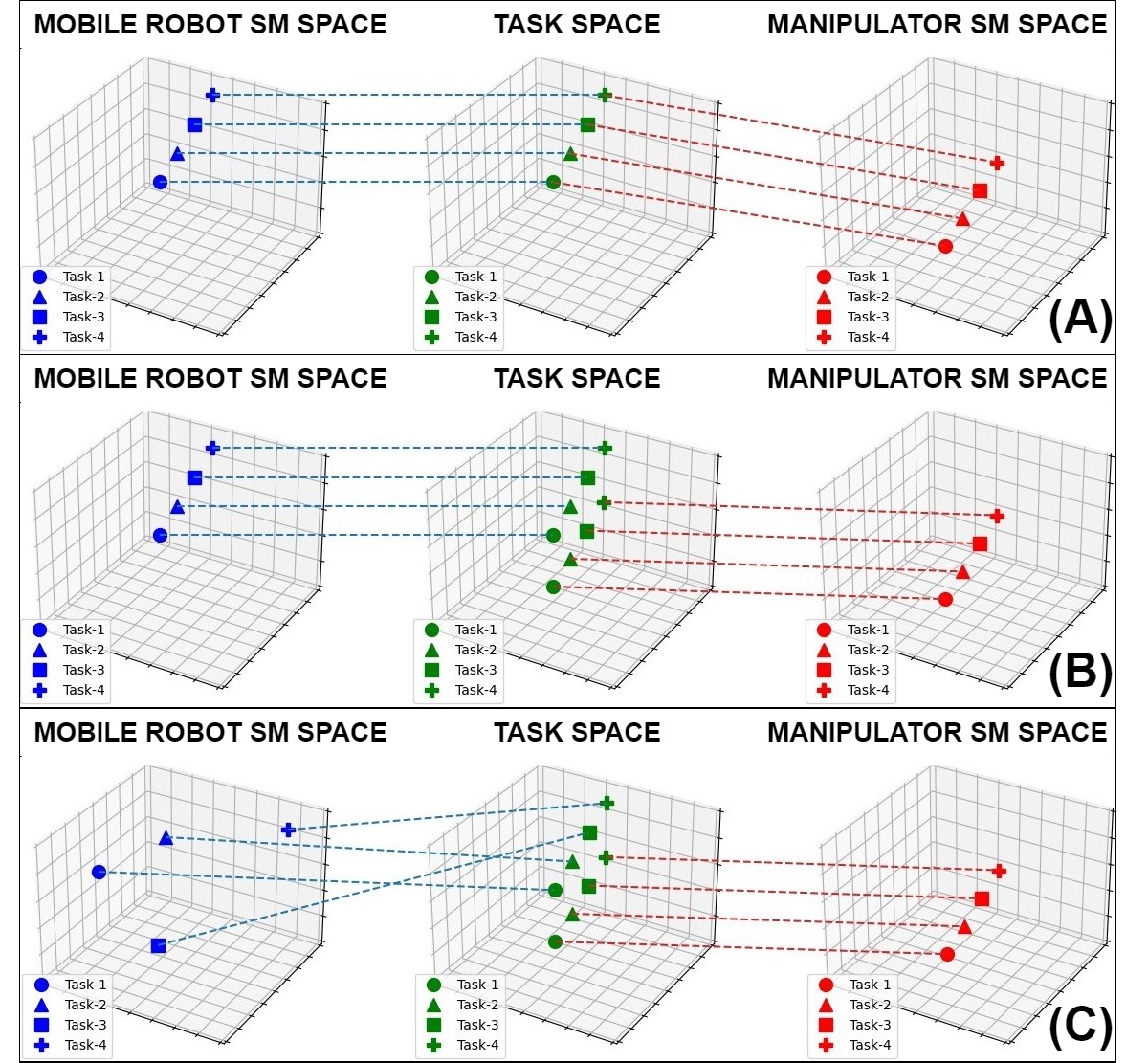}}
\caption{
The conceptual summary of the task-based correspondence learning problems studied in this paper. The middle, left, and right columns represent task and sensorimotor (SM) spaces of the mobile and manipulator robots, respectively. The shapes of the markers represent tasks. Each marker in the middle plot corresponds to the required robot execution trajectory in the task space to achieve the task. Each marker in the left and right plots represents the required SM trajectory of the corresponding robot to achieve the task. The dashed line between markers represents the mapping between SM and task spaces. In (A), the robots follow the same task-space trajectory to achieve the same task. In (B), the robots need to follow different trajectories to achieve the same task. In (C), the robots again need to follow different trajectories, but in this case, the complexity of the sensorimotor mappings is also different. 
}
\label{fig:sm-task-space}
\end{figure}

\erhan{In this work, we aim to learn the correspondences among robots with  different morphologies, including robots as different as  a fixed-based manipulator and a differential-drive mobile robot, that can learn to achieve the same tasks with significantly different bodies, actuators, and control variables.} 
Inspired by \cite{seker2022imitation}, our system forms a common latent representation between different robots. After learning the correspondences, and given observations of a new task execution from one robot, this common representation is used to generate the execution trajectory of the same task for the other robots. Each skill in each robot is encoded by a separate Conditional Neural Movement Primitives (CNMPs) network \cite{seker2019conditional}, which is an encoder-decoder network that can produce complex target trajectories from observations on these trajectories. Observations are given to the CNMP encoder to generate a latent representation for the target trajectory. The core idea is to blend the latent representations of each robot's CNMPs in order to form a common latent representation for both robots for the same task. This common latent representation is then fed to separate decoders of different robots to generate the execution trajectories of the corresponding robots. The core idea is that the coupled CNMP system is trained such that the common latent representation, which is able to generate execution trajectories for both robots, can be obtained even only from the observations (and therefore the latent encoding) of a single robot by setting the blending weights accordingly.
\erhan{We envision that this approach would be beneficial in real-life if  different robots are \emph{aligned}, possibly at production time, through the proposed method by using an initial fixed action set, such as manipulation of primitive shaped objects. Then, when one robot learns to perform a novel task, the other robots would be able to automatically have the ability to fulfill this task thanks to the initial alignment. From the viewpoint of world models \cite{taniguchi2023world}, our system’s objective can also be seen as using a correspondence between action trajectories in two or more models and forming a shared representation between them. }

%
%

A series of experiments are conducted to assess the extent of correspondence learning between the manipulator and mobile robots in tasks with different levels of complexity. For this, we designed these experiments to address the question of 
``Can task-specific correspondences between robots be learned''
\begin{itemize}
\item when the robots need to follow the same task-space trajectory to achieve the same task (Fig.~\ref{fig:sm-task-space}~(A))?
\item when the robots need to follow different trajectories to achieve the same task (Fig.~\ref{fig:sm-task-space}~(B))?
\item when the complexities of the required motor trajectories are different for different robots (Fig.~\ref{fig:sm-task-space}~(C))?

\end{itemize}

\begin{figure*}[t]
\centerline{\includegraphics[width=0.7\linewidth]{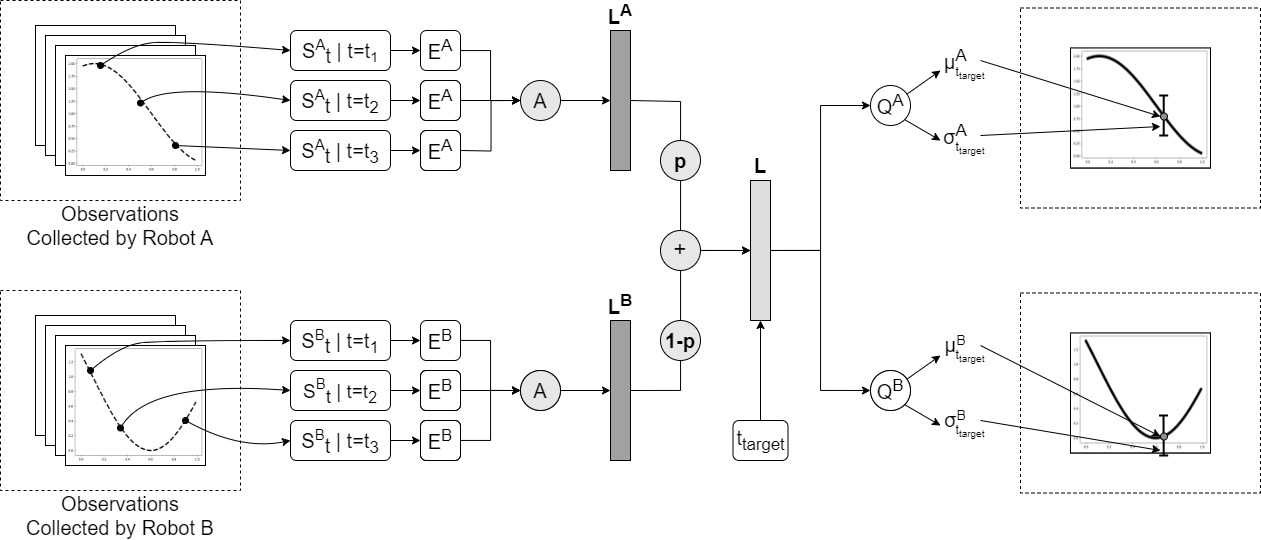}}
\caption{The general overview: Task-based correspondence via forming common latent representation between different robots.\hakan{ $E$ represents encoders, $S$ represents sampled observations, $L$ values represent latent representations and  $Q$ represents decoders}.  Given observations from an SM trajectory of one robot, our system can generate the full SM trajectory for the other robot to achieve the same task by setting the blending weight p to 0 or 1. Please refer to the text for the details of training and generation.}
\label{new-method}
\end{figure*}

\section{METHOD}

The general overview of our approach is given in Fig.~\ref{new-method}. Two (or possibly more) robots are provided initial proxy demonstrations for the same tasks. Depending on the embodiment and capabilities of the robots the demonstrations might differ in terms of the used modalities and the required trajectories. Given these demonstrations, the encoder networks (A) of the robots are trained together, forming a common latent space that can be used to generate the required trajectories for both robots. After training, a new task configuration can be demonstrated for one of the robots, either by providing an entire execution trajectory or a number of observations sampled from this trajectory. Using the common latent representation, our system can generate the full execution trajectory for the other robot to achieve the same task (B). We provide the details of the skill encoding, sharing, and generation in the rest of this section.

In our system, the skills are encoded and learned using CNMPs, and the latent representations of the CNMPs of different robots are blended to form a common latent representation. More formally,  let R = [$Robot_{1}, Robot_{2}, Robot_{3} ... , Robot_{n}$] be a set of sensorimotor information collected by n robots during action executions. The sensorimotor information during these executions is recorded at each time step. Let I be the multi-robot sensorimotor interaction data set, and the $j^{th}$ task execution is shown as $I_j = \{(t,S_{t}^R)\}_{t=0}^{T} $
where $S_{t}^{R}$ is the collection of sensorimotor data from all robots and t is time. $S_{t}^{R}$ is defined as $S_{t}^{R} = [S_{t}^{Robot_1}, S_{t}^{Robot_2}, S_{t}^{Robot_3}, ... ,S_{t}^{Robot_n} ]$, where each element holds the sensorimotor information collected by each robot. In this study, we used \hakan{several} robots whose morphologies and capabilities are different from each other to test the capabilities of our system. We aim  to learn correspondences \erhan{among the robots considered even when the means of achieving the target task may be significantly different for each robot. In the rest of the description, for the sake of brevity, we consider two robots, a manipulator and a mobile robot without loss of generality}.


Our proposed system connects multiple CNMP networks to find a common latent space. At the start of each training iteration, a task trajectory $I_d$ is sampled randomly from $I$. From $I$, an arbitrary number of data points of $(t,S_{t}^R)$  are sampled randomly as observations. 
The set of sampled observations is defined as $O^R = \{(t_i,S_{t_i}^R)\}_{i}^{obs_{max}} $ where $(t_i,S_{t_i}^R) \in I_d$ . \hakan{As an example}, we use two robots, a manipulator and a mobile robot. The observation set of these will be represented as $O^{manipulator}$ and $O^{mobile}$.  Other than $O^R$, a target tuple $(t_{target},S_{t_{target}}^R)$ is also sampled from $I_d$ which is used to learn the distributions on $t_{target}$ for observations collected by all robots.

The purpose of the model is to find a common latent representation between observations of all robots. To this end, the observations of each robot $O^r$ are converted to latent representations, which are calculated using the following equation: 

\begin{equation} \label{eq:1}
L_{i}^r = E^{r}((t_i,S_{t_i}^r)|\theta^r) \quad (t_i,S_{t_i}^r) \in O^r, r \in R
\end{equation}

where $E^r$ is a deep encoder for robot $r$ with weights $\theta^r$, and $L_i^r$ is the latent representation constructed using the given observation. After the encoding process, we obtain $L_i^{manipulator}$ and $L_i^{mobile}$  for the manipulator and the mobile robot, respectively. Following the construction of  $L_i^r$'s, these representations are averaged for each robot: 
\begin{equation} \label{eq:2}
L^r = \frac{1}{n} \sum_{i}^{n} L_{i}^r \quad r \in R
\end{equation}
where $n$ is the number of sampled observations during this iteration. Following this equation, we calculate  $L^{manipulator}$ and $L^{mobile}$, which represent the latent representations constructed using $O^{manipulator}$ and $O^{mobile}$, respectively. We aim to find a common latent representation using the constructed latent representations to generate the required trajectories for both robots. To accomplish this, a general latent representation $L$ is constructed by taking the convex combination of all $L^r$:
\begin{equation} \label{eq:3}
L = \sum_{r}^{R} p^r L^r \quad 0\le p^r \le 1, \sum p^r = 1, r \in R
\end{equation}

This calculation is applied to blend all latent representations into one common latent space.
\hakan{Each $p^r$ can be used to control the information flow from its corresponding encoder $E^r$. For instance, if $p^k$ is close to zero, the effect of $L^k$ constructed by $E^k$ in the formation of $L$ is low. Similarly, if $p^k$ is close to 1, its effect is high. During training, the $p$ values are uniformly sampled between zero and one and then normalized, i.e. $L$ is set to a (uniformly) random convex combination of $L^r$s, which is computed at each gradient update step.}

After all representations are merged into one, this representation is decoded to obtain target distributions on $t_{target}$ for all robots: 
\begin{equation} \label{eq:4}
(\mu_{t_{target}}^r,\sigma_{t_{target}}^r ) = Q^{r}((L,t_{target})|\phi^r) \quad r \in R
\end{equation}
where $Q^r$ is a deep decoder with weights $\phi^r$ that constructs distributions with mean $\mu_{t_{target}}^r$ and variance $\sigma_{t_{target}}^r$ for the robot $r$. For our robots, 
 $(\mu_{t_{target}}^{manipulator},\sigma_{t_{target}}^{manipulator} )$ and  $(\mu_{t_{target}}^{mobile},\sigma_{t_{target}}^{mobile} )$ are obtained using $Q^{manipulator}$ and $Q^{mobile}$, respectively. The learning goal of our system is to obtain more accurate distributions for given condition point(s) and the target point, and therefore the loss is defined similarly to \cite{seker2019conditional}:
\begin{equation} \label{eq:5}
Loss = -\sum_{r}^{R}  \log P(S_{t_{target}^r}|\mu_{t_{target}}^r,\sigma_{t_{target}}^r ) \quad r \in R
\end{equation}
 where $S_{t_{target}}^r \in S_{t_{target}}^R$ is the target SM state for agent $r$. \hakan{During training, deep decoders $Q$ and  encoders $E$ are trained. At each training step, $O^R$ and corresponding $p^r$ values are given to the network as inputs along with a $t_{target}$ value. The loss is calculated using the network's prediction and  $S_{t_{target}}^R$.}

 After sufficient training, the system can make action predictions for all time steps of that action for all robots when a distinct $O^R$ is provided to the system. More importantly, when querying, trajectory generation can be achieved by making $p^r$ for \hakan{the source robot} \hakan{one}. \hakan{This makes the rest of the $p$ values zero since we are using convex combination. Hence, only $L^r$ is used to form $L$ which is then used to construct the trajectory outputs by all decoders.} \erhan{We also experimented with binary $p$ values reminiscent of a dropout \cite{srivastava2014dropout} like idea, which was also effective in learning the correspondences, albeit at a slower rate. A notable difference we observed is that,  a common latent space was not formed, and the representations constructed by the encoders were sparse when binary $p$s were used. Thus, in the experiments reported continuous $p$ values were used. The results showed that, our system was able to generate the desired trajectories for  all the involved robots when only one robot acted as the \emph{source} or \emph{teacher} by providing the demonstrations for the novel task.}
 

\begin{figure}[t]
\centerline{\includegraphics[width=.7\linewidth]{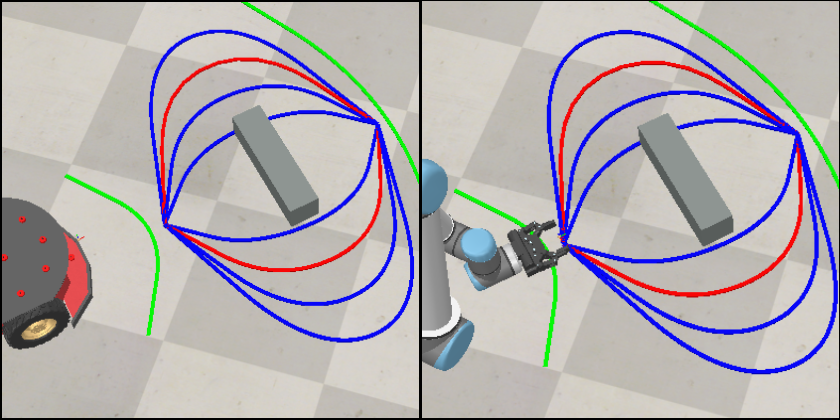}}
\caption{
Training and test paths correspond to the paths required to achieve the obstacle avoidance tasks in Section \ref{A}. Left and right figures show the manipulator and mobile robots. Blue and red paths were used for training and testing, respectively. Green lines show the action range of the manipulator. 
}
\label{object-avoidance-paths}
\end{figure}
\begin{figure}[b]
\centerline{\includegraphics[width=0.85\linewidth]{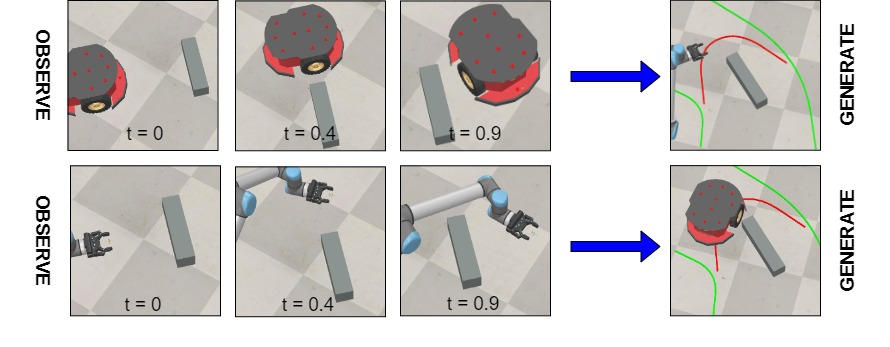}}
\caption{The snapshots from the mobile robot to the manipulator (on top) and from the manipulator to the mobile robot (on bottom) for the  experiment in Section \ref{A}.  Using three observations from the given trajectory on a new task configuration (on the left of the figure), the system is able to generate the desired trajectories for the other robot (on the right ).}
\label{object-avoidance-generation}
\end{figure}
\begin{figure}[t]
\centerline{\includegraphics[width=0.75\linewidth]{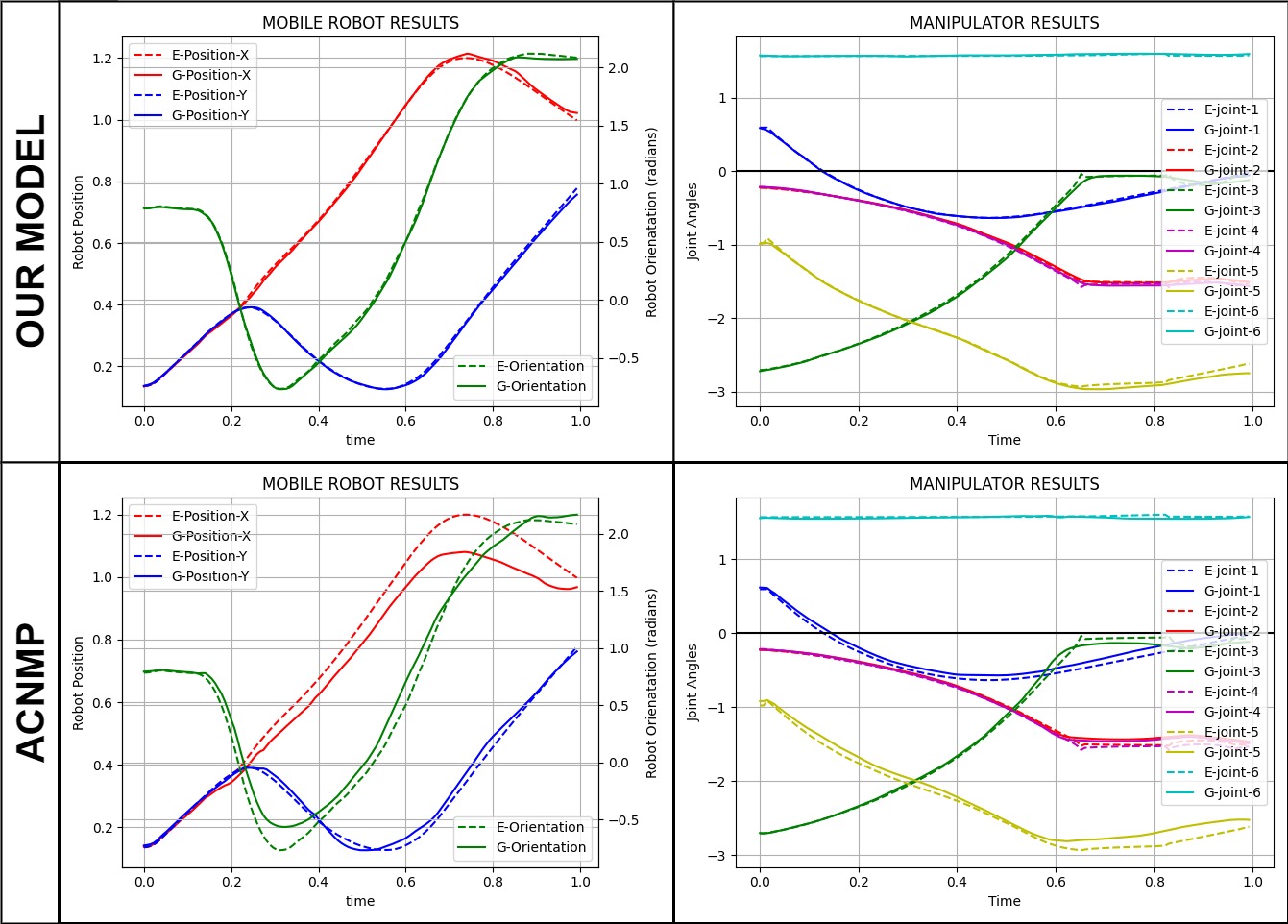}}
\caption{Generated trajectories of one of the data sets of  experiment in Section \ref{A} with their ground truth values.  Dashed lines are expected ground truth values and solid lines represent the generated values. The upper part shows the results obtained using our model while the lower part shows the results obtained using the ACNMP approach. }
\label{object-avoidance-results}
\end{figure}

\section{EXPERIMENTS}
To analyze the capabilities and limits of our method, we designed three experiments (Fig.\ref{fig:sm-task-space}). \hakan{We used a mobile robot (Pioneer3-DX), a manipulator with 6 degrees of freedom (UR-10), and a manipulator with 7 DOF named (Kuka LBR4+). We also used 3 different grippers for the manipulators namely Robotiq85, BarrettHand, and Robotiq 3-finger adaptive gripper. The last one is used in the real-world experiment. We used 3 manipulators with different grippers to show the practicality of our model, because although they are morphologically different robots, most of the tasks they can perform are shared between them. Alongside them, we chose to use a mobile robot to test whether our model can perform when there is a significant difference in the robot morphologies.} The mobile robot is controlled by setting the desired position and orientation of the robot body at each step. Therefore $S^{mobile}$ includes position and orientation. The manipulators are controlled by setting the desired joint angles of the robot arms, therefore  $S^{manipulator}$ includes the joint angles.  The simulation experiments were conducted in the CoppeliaSim\cite{rohmer2013v}.

\subsection{Correspondence learning when the task requires the same Cartesian path}\label{A}



\hakan{This experiment, where Pioneer 3D-X and UR-10 with Robotiq85 gripper were used, aims to verify that our method can learn the correspondences between two robots  when they need to follow the same path in task space.} An obstacle avoidance task is designed where the robots are required to avoid the obstacle using different paths. 8 different obstacle avoidance trajectories were collected for each robot as shown in Fig.~\ref{object-avoidance-paths} where 6 trajectories were used for training and the rest were used for testing.  
After paired training is completed with a number of demonstration trajectories, where position and orientation are used for the mobile robot and joint angles are used for the manipulator robot, the model was queried to construct the trajectories for a robot given a number of the observation points from the other robot at certain time steps as input. When conditioned with an arbitrary number of points collected by any robot, the system was able to construct the corresponding trajectories. In other words, the manipulator/mobile robot was able to avoid the obstacles in the test and training setups following the trajectories generated based on task input from the other robot.

Fig.~\ref{object-avoidance-generation} provides a number of snapshots for task configurations that were not observed during training. On the top and bottom rows,  transfers from mobile robot to manipulator and from manipulator to mobile robot are shown, respectively. On the top left, the mobile robot is provided with a new movement trajectory that was not seen during training. Our system uses observations (position and orientation of the mobile robot) from this trajectory at three different time steps and is able to generate the desired movement trajectory (joint angles) for the manipulator robot, whose corresponding motion is shown on the right.

We analyzed the results by comparing the trajectories generated by the model and the desired trajectories. We applied this process both to training trajectories and test trajectories. For both cases, we obtained results with negligible error. Generated trajectories of one of the test cases can be seen in the upper part Fig. \ref{object-avoidance-results} with the desired trajectories.
\begin{figure}[t]
\centerline{\includegraphics[width=0.75\linewidth]{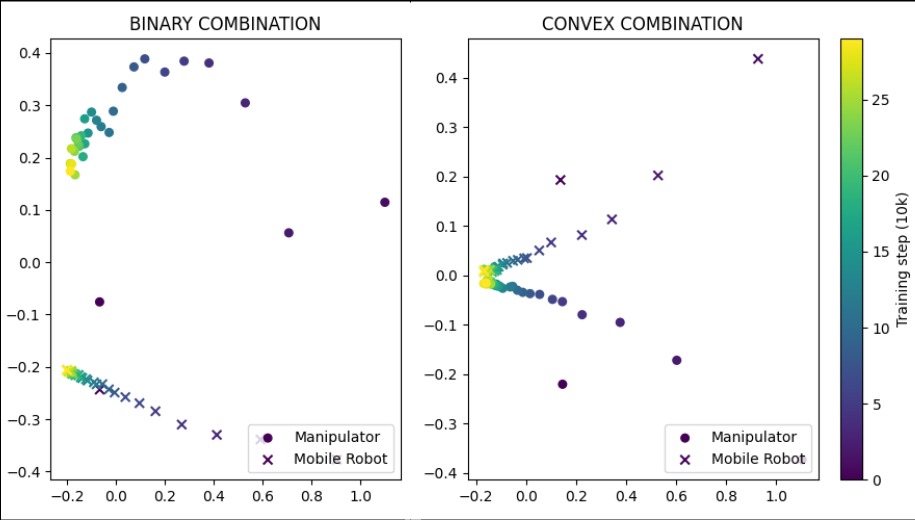}}
\caption{The dimensionality analysis of the representations constructed by the encoders when the model is trained using binary and convex combination. While representations formed using convex combination converge as the training progresses, the representations formed using binary combination do not.}
\label{EXP1-latent}
\end{figure}

\hakan{We also compared our method with a baseline method \cite{akbulut2021acnmp} where an additional loss function was realized as the mean squared error between $L^r$ values of different CNMP networks. This loss function also needed to set the magnitude of this value as a hyper-parameter.
This value must be adjusted with caution because while using a small magnitude is not enough to force common representations to be formed, using a large magnitude causes the additional loss to dominate the main loss function used by the CNMPs. Hence, the magnitude value has to be carefully optimized for this approach to work properly while our approach does not require any additional optimization. Furthermore, despite the optimization, our model is more accurate and converges faster. The results for the same test case can be seen in \ref{object-avoidance-results}. The results are the best results we could obtain for each model. Procuring these results took 200k training steps with our model while it took 300k steps to train the ACNMP model\erhan{, with approximately the same number of learnable parameters in each model. \hakan{During testing, the total mean squared trajectory error of ACNMP was almost twice of the error of our model both for mobile (3.426 vs.6.430)  and joint (2.971 vs. 6.841)  outputs.
} Finally, each training step of ACNMP is more computationally heavy since it involves an additional computation for the additional loss function.}
}

\hakan{As mentioned in the Method section,  we also experimented using binary combination instead of convex combination. Although we did not observe any significant decrease in accuracy between the two, analyzing the latent representations formed by the encoders showed that the system was unable to find a common latent space when binary combination is used. The difference between them can be seen in Figure \ref{EXP1-latent}. We picked a couple of novel points along one of the test trajectories and constructed latent representations of each robot using their encoders. We used Principal Component Analysis to decrease the dimension of each latent representation to two and plotted how the representations change as the training progresses. It can be seen on the figure that when convex combination is used as training progresses the latent representations of the two robots converge to a common representation. However, this is not the case when binary combination is used.}

\begin{figure}[]
\centerline{\includegraphics[width=0.7\linewidth]{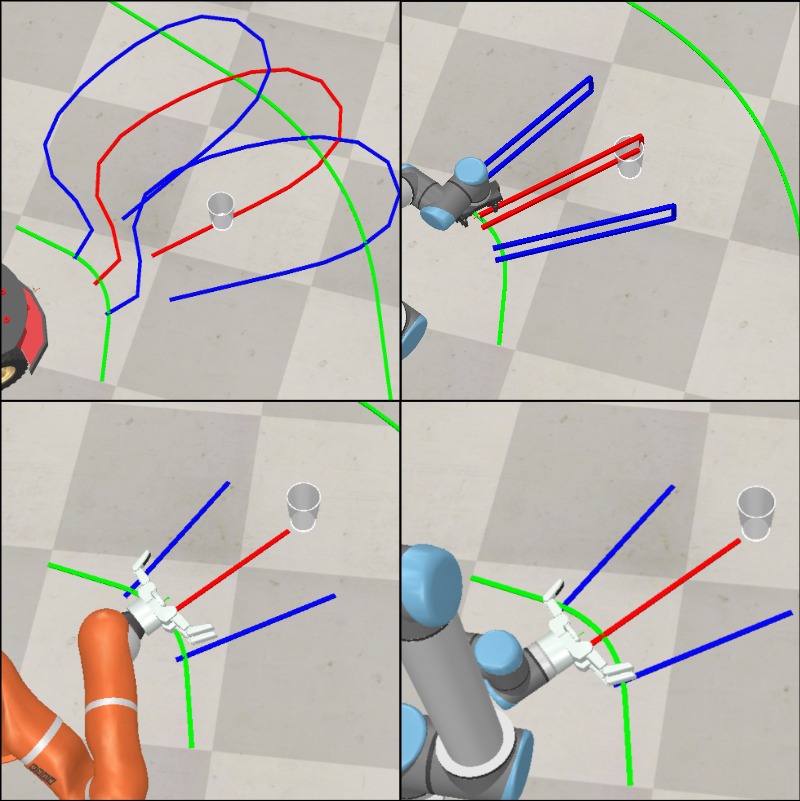}}
\caption{A number of sample demonstration paths collected for cup retrieval tasks of the experiment in Section \ref{C}. The blue paths are for training and the red paths are for testing. Green lines represent the action range of the manipulator.}
\label{cup-pulling-paths}
\end{figure}




\subsection{Correspondence learning when different Cartesian paths are needed for task completion}\label{C}

The morphologies and the capabilities of \hakan{different robots} are dissimilar, and therefore they can achieve the same task using different means depending on the task and the involved objects. 
In this section, we aim to test whether our system can learn the correspondence between robots even if the means to achieve the same task, i.e. the paths they follow, are different (Fig.~\ref{fig:sm-task-space} (B)). \hakan{For this purpose, we used the mobile robot alongside with UR-10 and KUKA LBR4+ to also show the scalability of our model. We used UR-10 with 2 different grippers namely Robotiq85 and BarrettHand while only using the latter with the KUKA LBR4+. We discretized the gripper status as open and closed while using it as input to the system.} As a use case, we selected a task that required the robots to retrieve a cup to a desired location. In this setup, while the demonstrator used one of the fingers of the \hakan{Robotiq85} gripper of the \hakan{UR-10} manipulator as a hook to pull the cup to itself, the mobile robot is driven behind the cup to push it to the same location. \hakan{The ones with the BarrettHand performed the same action realized by approaching the cup with the open gripper, grasping the object by closing the gripper, and pulling the cup towards the robot respectively.} The location of the cup is changed, while the performed strategy remains the same. Note that the cup is only chosen to represent the goal and no information about the cup is used by either of the robots or by any part of our model.  To show the generalization capability of our system, during testing, the cup is placed at a position that was not observed during training. Seven demonstrations were collected for each robot. In order to provide a clear picture, we presented only three of them in Fig.~\ref{cup-pulling-paths}. \hakan{The generation process is similar to the previous section.}



\begin{figure}[t]
\centerline{\includegraphics[width=0.75\linewidth]{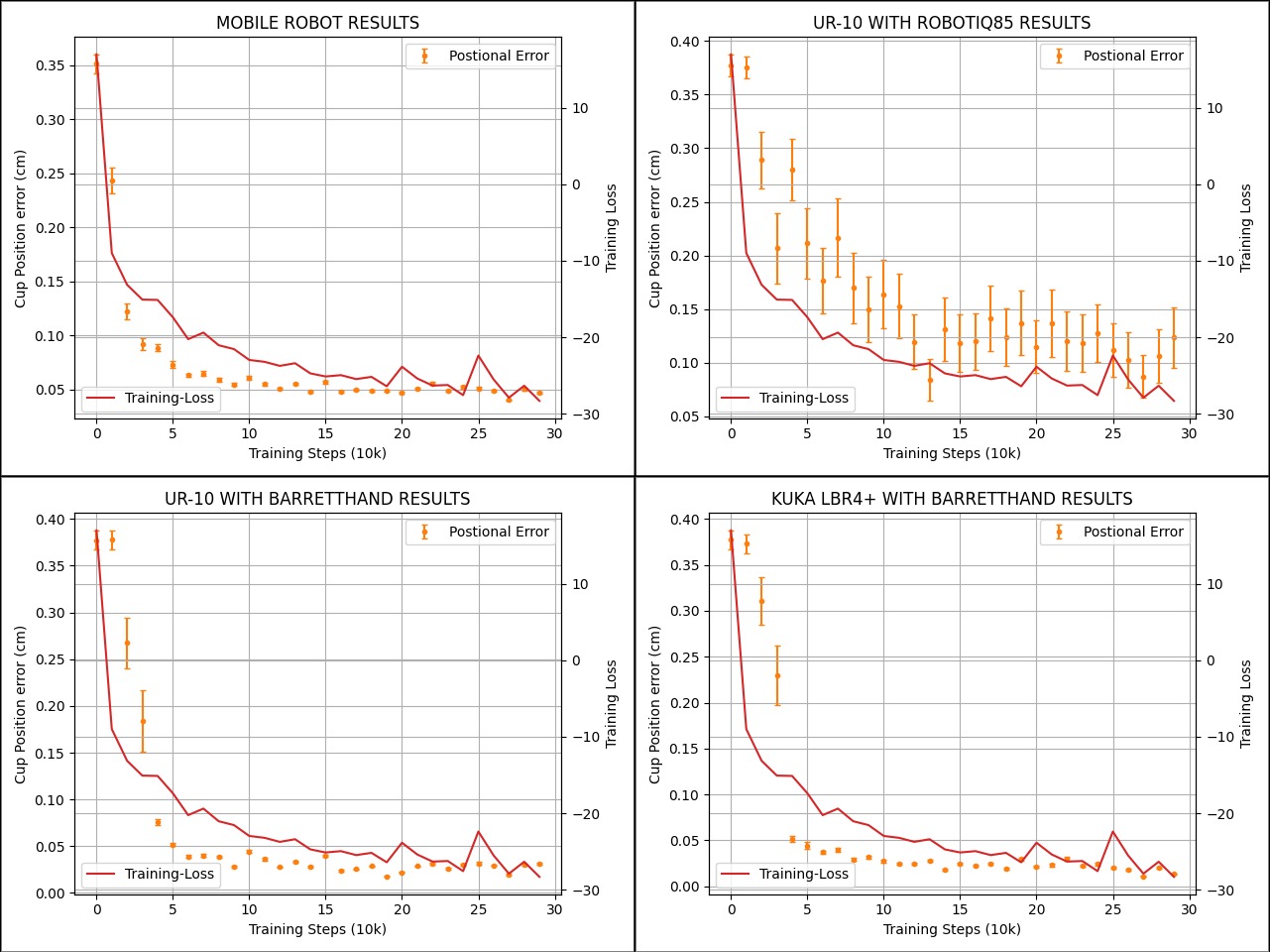}}
\caption{The results of the cup retrieval tasks in the correspondence learning when different Cartesian paths are needed for task completion experiment in section \ref{C}. Graphs show how as training progresses and training loss decreases, the distance of the cup from the desired position changes. Observations from the same time steps are used in all runs (First, thirtieth, sixtieth, and the last points of a sample with length 128).}
\label{cup-pulling-results}
\end{figure}

This experiment was designed to retrieve the cup to a goal position. Therefore the task success can be evaluated by comparing the goal position and the final position of the object after action execution (even though no information related to the object is used during control). We measured the error between the desired and actual positions for the transfer in both directions at different time steps during training and plot the error statistics along with the training loss in Fig.~\ref{cup-pulling-results}. First of all, the results provide the mean and variance of the final position errors obtained with 30 runs in the novel task configurations. As shown, the final position error of the object drops with training and consistently with the decrease in the loss of our model. \hakan{The results were similar regardless of the robot chosen as input (the one with $p=1$ during test time), so we chose to average the values to show the results more clearly.  The fact that there was no notable difference shows that our system achieved generalization in transfer in the interpolation cases. Across all plots, it can be seen that the positional error values drop as the training loss decreases, hence as the system learns, the success rate increases. The mean and the variance values of manipulators with BarrettHand grippers decreased more slowly than the mobile robot's because in the early stages of training, the gripper mostly fails to grasp the cup, but after a while, it never misses it which resulted in slightly less positional error values than the mobile robot's results. This is because the gripper does not allow the cup to slide while the mobile robot lacks the capability to keep the cup in place. Furthermore, we did not observe any significant difference in results between the UR-10 and KUKA LBR4+ when they both use BarrettHand as gripper which shows that the degree of freedom of the manipulator used does not affect the model's performance significantly. However, the mean and the variance values of UR-10 with ROBOTIQ85 gripper were higher than all others.}Upon observing this behavior, we noticed that minor deviations in the manipulator's trajectory could lead to failure to insert the finger into the cup, thereby making it impossible to move the cup to the desired position. This failure is not attributed to our correspondence learning system because the same failure happens when the same robot is used for observation and generation.


\begin{figure}[t]
\centerline{\includegraphics[width=0.6\linewidth]{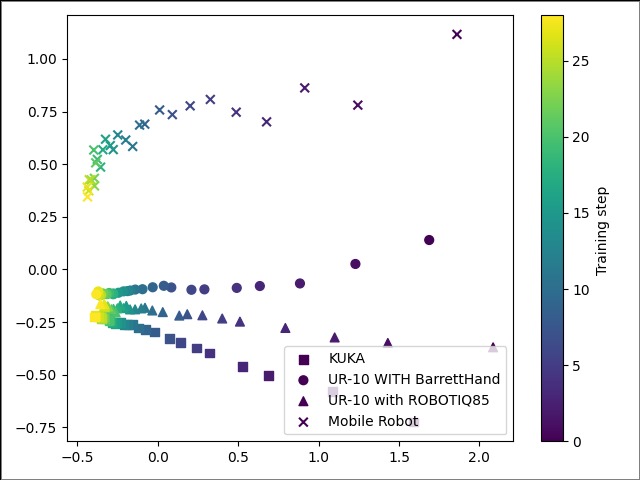}}
\caption{The dimensionality analysis of the representations constructed by the encoders of different robots. It can be seen from the figure that although all representations come closer than the initial point, representations of morphologically less different robots converge closer.}
\label{EXP2-latent}
\end{figure}


\begin{figure}[b]
\centerline{\includegraphics[width=.7\linewidth]{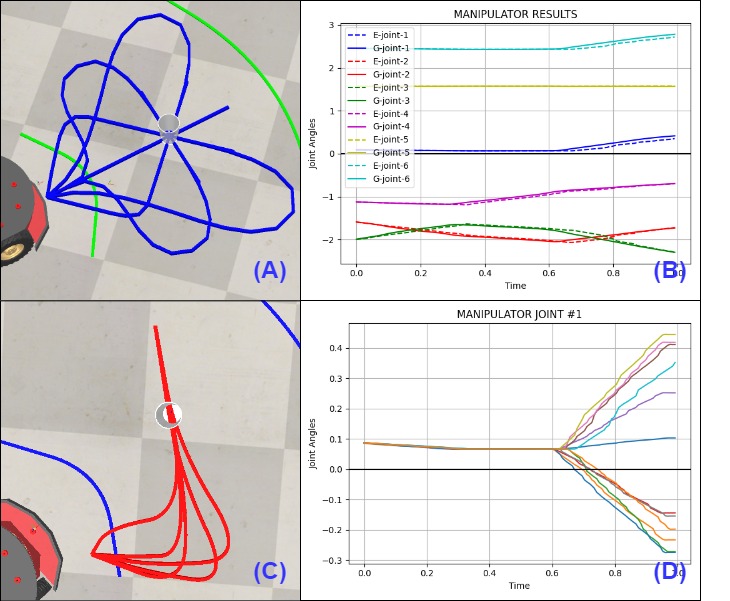}}
\caption{(A) Training set used in cup retrieval tasks of experiment in Section \ref{D}. (B) One of the test results for the manipulator in the same experiment. (C) A number of the paths the system may choose to generalize in the same experiment. (D) All trajectories used in training and testing of one of the joints of the manipulator in the same experiment.}
\label{fig:real-robot-results}
\end{figure}
\hakan{We compared our model with the ACNMP approach in this experiment from the perspective of scalability. While our model can be scaled to any number of robots by increasing the number of encoder-decoder couples, this is not the case for the ACNMP. For every additional robot, additional loss functions that take the mean squared error between the new robot's latent representation and the rest of the representations have to be added. For instance, for four, five, and six robots this number becomes six, ten, and fifteen respectively. Consequently, as the number of robots increases, the computation required to train the model increases factorially. Additionally, each additional loss introduces a new hyper-parameter to optimize separately.
In our attempts, after significant optimization effort, we were able to obtain results similar to the ones we obtained in Figure \ref{object-avoidance-results} regarding accuracy with significantly higher training time.}

\hakan{We analyzed how the latent representations formed by the encoders change over the course of training in this experiment as well to show whether using different number of robots with different morphology differences affect the latent space formation. We used the same process we used in Section \ref{A} for dimensionality reduction. Analysis results can be seen in Figure \ref{EXP2-latent}. It can be seen that despite the increase in the number of robots, our model can still find a common latent space between them. However, the difference in morphology seems to affect this formation. While representations of all robots become closer to each other than they were initially, the robots with less morphology difference (manipulators) converge closer. Nonetheless, the difference is not very large and did not cause any notable decrease in accuracy.}

\begin{figure}[t]
\centerline{\includegraphics[width=0.9\linewidth]{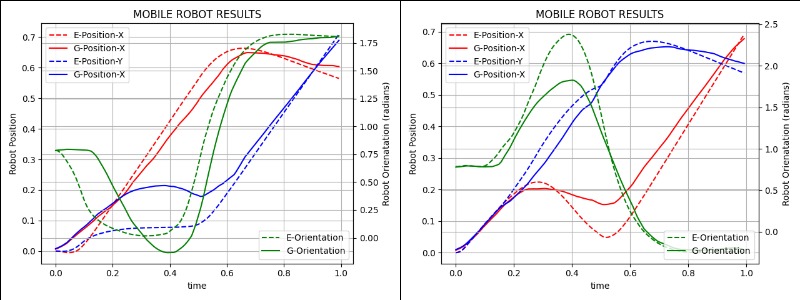}}
\caption{Mobile robot test results of the experiment in Section \ref{D}. Dashed lines are expected ground truth values and solid lines represent the generated values.}
\label{real-mobile-robot-results}
\end{figure}



\subsection{Correspondence learning when task completion requires different levels of complexity in the SM spaces}\label{D}
\begin{table}[b]
\caption{Directional and positional error values of all test cases of cup retrieval tasks of experiment in Section \ref{D} for the mobile robot. Dir. error is the difference between the angle the mobile robot pushed the cup to and the desired angle for that case. Pos. error is the difference between the desired and actual final position of the cup.}
\begin{center}
\begin{tabular}{|c|c|c|c|c|}
\hline
\textbf{Test}&\multicolumn{4}{|c|}{\textbf{Positional and Angular Error}} \\
\cline{2-5} 
\textbf{Case} & \textbf{\textit{Desired Angle}}& \textbf{\textit{Outcome}}& \textbf{\textit{Directional Error}}& \textbf{\textit{Pos. Error}} \\
\hline
1& 30& 26.739&3.261 &0.0226 \\
\hline
2& -60&-62.082 &2.082 & 0.0587\\
\hline
3& -30&-29.646 &0.354 &0.0514 \\
\hline
4& 60& 63.176&3.176 &0.0514 \\
\hline
\multicolumn{4}{l}{}
\end{tabular}
\label{real-robot-table}
\end{center}
\end{table}



In the previous experiment, the robots needed to follow different paths in order to achieve the same task, and gradual changes in the sensorimotor spaces of both robots were required for the gradual change in the task space. In this section, we aim to test our system when the trajectory of one robot needs to change significantly in order to achieve gradually changing  task configurations (Fig.\ref{fig:sm-task-space} (C)). In order to address this challenge, we updated the cup retrieval task by keeping the cup in the same position and moving it to different goal positions by pulling/pushing the cup in different directions (Fig.\ref{fig:real-robot-results}). This way, we managed to increase the level of complexity of the actions of the mobile robot while decreasing it for the manipulator. This difference in complexity can be seen in panels (A) and (D) of Fig.\ref{fig:real-robot-results}. Panel (A) shows the paths followed by the mobile robot to move the cup to the desired position, and  panel (D) shows the joint angle trajectories for different desired positions. As shown while there is a gradual change in the SM trajectory of the manipulator, the change in the SM of the mobile robot is very complex. Additionally, only the last parts of the joint angle trajectories are different for all trajectories of all joints. The identical part represents when the gripper is moving towards the cup and putting its finger in the cup, and the rest represents pulling the cup to different angles. We used seven trajectories for training and four trajectories for testing for this section. Another objective of this experiment was to present a proof-of-the-concept realization in the real world. Therefore, we used a real manipulator, a real UR-10 with a 3-finger gripper.

\begin{figure}[t!]
\centerline{\includegraphics[width=0.85\linewidth]{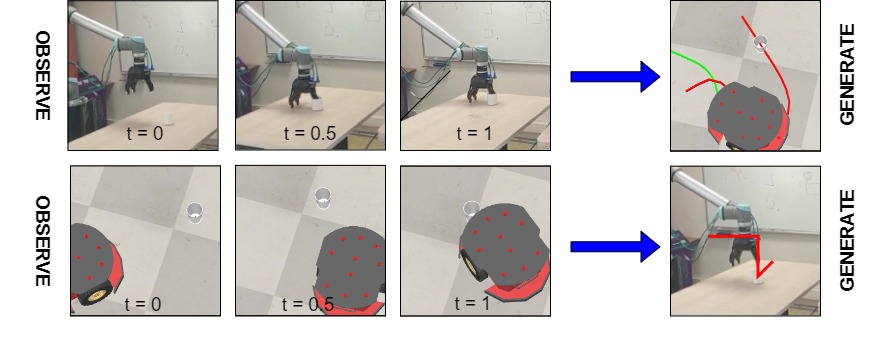}}
\caption{Snapshots from mobile robot to manipulator (on top) and from manipulator to mobile robot (on bottom) for the experiment in Section \ref{D}.  Using three observations from the given trajectory on a new task (on the left of the figure), the system is able to generate the desired trajectories for the other robot (on the right ).}
\label{real-robot-generation}
\end{figure}


We observed that the difference in complexity created  uncertainty in the trajectory level for different task configurations between robots. The reason is that the trajectories used to approach the cup until making contact with it are the same for the manipulator independent of where the cup is being moved, whereas the approach trajectory depends on the final goal position for the mobile robot. We observed this affected the way the system generalizes for the test cases. While the system can generate the desired trajectories close to how we envisioned them before testing (Fig.~\ref{fig:real-robot-results} (B)) for the manipulator, this was not the case for the mobile robot (Fig. \ref{real-mobile-robot-results}). However, while there are significant shifts between the desired trajectories and the generated ones in Fig. \ref{real-mobile-robot-results}, it can be seen that the last parts of the trajectories are close to the desired ones. This can be explained by the uncertainty. While there is only one way for the manipulator to achieve the task (with the same approach trajectory), there are numerous ways for the mobile robot (with slightly different approach trajectories). As seen in Fig.~\ref{fig:real-robot-results} (C), to push the cup to the desired location, the system can generalize to any of the red paths shown and to many others. It can be seen in Table~\ref{real-robot-table} that although the system is unable to generate the trajectories for the mobile robot as we predicted them, it can successfully achieve the given task with small errors.

\section{CONCLUSION}

In this work, we proposed and realized  a correspondence learning framework that enables task-specific skill transfer between robots with different morphologies and allows goal-based imitation with different means and robots. 
Our system was able to learn the task-level correspondence between the pose space of a differential drive mobile robot and the joint spaces of fixed-based manipulators. The robots have different bodies and sensorimotor spaces and may need to follow different paths to achieve the same task, yet, our system successfully learned this correspondence. 
Even though our system was shown to enable task-level skill transfer between manipulation and mobile robots in cup retrieving tasks that require different movement trajectories, the geometric or visual properties of the interacted objects were not used by our system to make generalizations across tasks.  In the future, we plan to also focus on using object features, such as object images or high-level features, in order to transfer and generalize object affordances \cite{cakmak2007affordances} among different robots. \hakan{We also plan to experiment with different modalities of each robot such as the motor velocities of mobile robots or joint velocities of manipulators. Although the velocity alone cannot be used solely to prescribe a path in space that can be used to accomplish a task, we believe by using positional information from earlier steps the future versions of our system can infer paths in space. In order to see how our model deals with large sensorimotor dimension differences, experiments with various agents such as humanoids, musculoskeletal robots, and even humans can be conducted, and we plan to investigate this in a future work. Finally, we plan to investigate if skill transfer is possible using our model.}


\section*{ACKNOWLEDGMENT}

This research has been funded by the JST Moonshot R\&D, Japan (JPMJMS2292), by the JST CREST, Japan (JPMJCR21P4), by the JSPS KAKENHI, Japan (21H05053), by the World Premier International Research Center Initiative (WPI), MEXT, Japan, by Japan Society for the Promotion of Science, Grant-in-Aid for Scientific Research – the project (22H03670), the project JPNP16007 commissioned by the New Energy and Industrial Technology Development Organization (NEDO), and the Scientific and Technological Research Council of Turkey (TUBITAK, 118E923). The authors would like to thank Muhammet Hatipoglu for his help in the real robot experiments.

 \bibliographystyle{IEEEtran}
 \bibliography{ref}

\section*{APPENDIX}
For the deep encoders and decoders of the network fully connected layers with ReLU activation functions are used. For instance the sizes of layers used in section \ref{A} are as follows:
\begin{itemize}
    \item Mobile robot encoder : Input, 32, 64, 64 ,128, 256, 128
    \item  Manipulator encoder : Input, 32, 64, 64 ,128, 256, 128
    \item   Mobile robot decoder : 512, 216, 128, 32, Output
    \item Manipulator decoder : 512, 216, 128, 32, Output
\end{itemize}


\end{document}